\documentclass[letterpaper, 10 pt, conference]{ieeeconf}  

\IEEEoverridecommandlockouts                              

\usepackage{graphics} 
\usepackage{epsfig} 
\usepackage{mathptmx} 
\usepackage{times} 
\usepackage{amsmath} 
\usepackage{amssymb}  
\usepackage{longtable}
\usepackage{subcaption}
\usepackage[noadjust]{cite}
\usepackage{balance}
\usepackage{xcolor}
\usepackage[capitalise]{cleveref}

\definecolor{roboat}{RGB}{255, 128, 0}
\definecolor{obstacle}{RGB}{0, 200, 0}
\definecolor{obstacleblue}{RGB}{102 102 255}
\definecolor{roboatyellow}{RGB}{210 210 0}
\definecolor{refpath}{RGB}{0 0 255}

\title{\LARGE \bf
Regulations Aware Motion Planning for Autonomous Surface Vessels in Urban Canals
}

\author{Jitske de Vries, Elia Trevisan, Jules van der Toorn, Tuhin Das, Bruno Brito, and Javier Alonso-Mora 
\thanks{This work was supported by the TRiLOGy project and the Amsterdam Institute for Advanced Metropolitan Solutions (AMS) in the Netherlands.}
\thanks{The authors are with the Department of Cognitive Robotics, Delft University of Technology, 2628 CD, Delft, The Netherlands.
        {\tt\small \{e.trevisan; bruno.debrito; j.alonsomora\}@tudelft.nl}}%
}

\begin{document}

\maketitle
\thispagestyle{empty}
\pagestyle{empty}

\begin{abstract}
In unstructured urban canals, regulation-aware interactions with other vessels are essential for collision avoidance and social compliance.
In this paper, we propose a regulations aware motion planning framework for Autonomous Surface Vessels (ASVs) that accounts for dynamic and static obstacles. Our method builds upon local model predictive contouring control (LMPCC) to generate motion plans satisfying kino-dynamic and collision constraints in real-time while including regulation awareness. To incorporate regulations in the planning stage, we propose a cost function encouraging compliance with rules describing interactions with other vessels similar to COLlision avoidance REGulations at sea (COLREGs). These regulations are essential to make an ASV behave in a predictable and socially compliant manner with regard to other vessels.
We compare the framework against baseline methods and show more effective regulation-compliance avoidance of moving obstacles with our motion planner. Additionally, we present experimental results in an outdoor environment.
\end{abstract}

\section{INTRODUCTION}
With a growing number of citizens and tourists, the scarce public space, roads, and public transport in Amsterdam is experiencing rising pressure \cite{GemeenteAmsterdam2021AmsterdamseBereikbaarheid}. A possible solution to this problem is to use the 165 canals with a total length of 100 km as an alternative to the conventional routes to transport goods and people. This opens up the opportunity for developing Autonomous Surface Vessels (ASVs) explicitly designed for urban environments, such as Roboat \cite{Wang}. 

\begin{figure}[tp]
  \centering
  \includegraphics[width=0.8\linewidth]{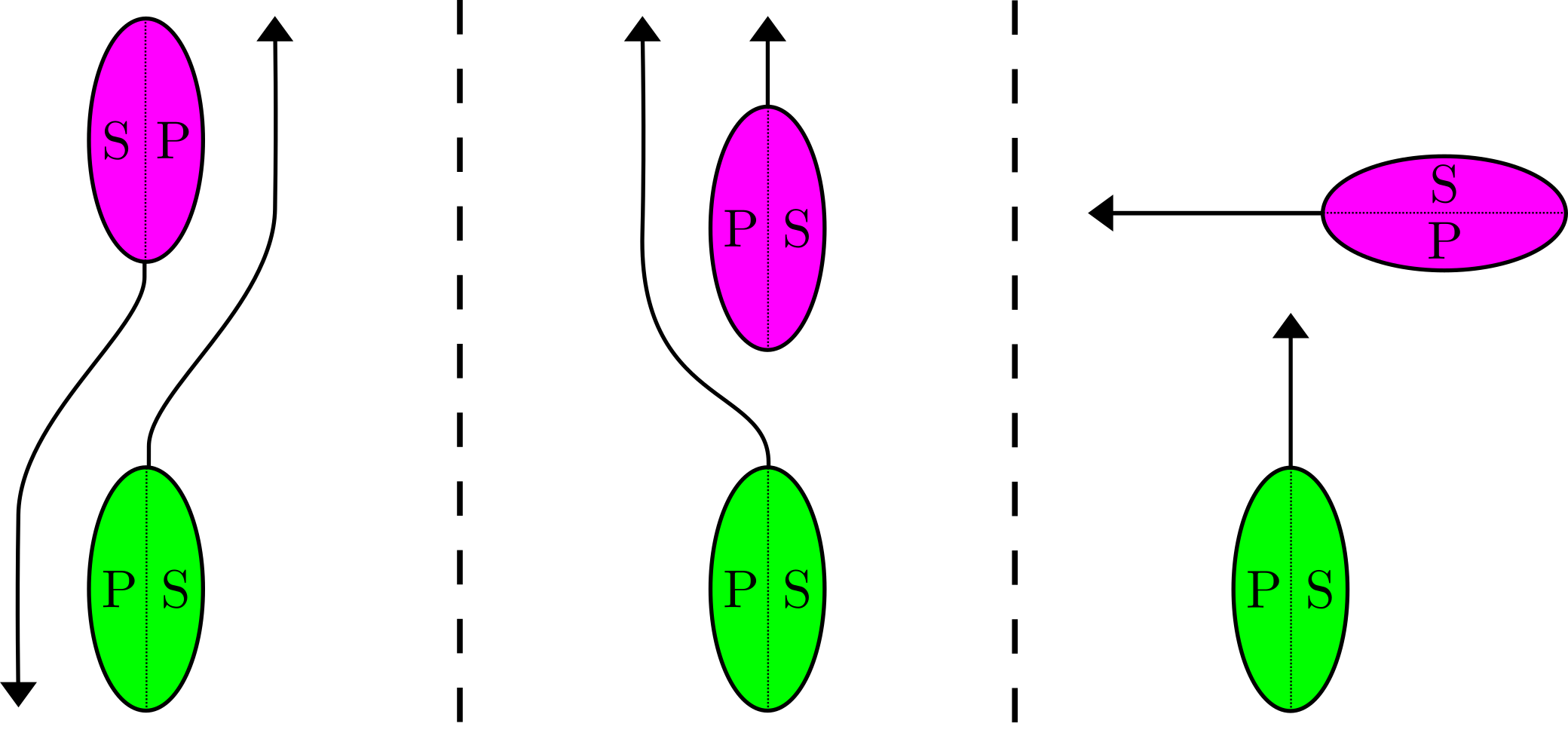}
  \caption{Two boats interacting in a (from left to right) head-on, overtaking, and crossing scenario according to the regulations. The starboard and port side of the boats are denoted as respectively S and P.}
  \label{fig:regulations}
\end{figure}

However, urban canals are challenging for motion planning since the space can be narrow and contain other human-operated boats. Also, ASVs have slow dynamics which lead to limited agility. Therefore, navigation requires precision and planning ahead to avoid any collision with both static and dynamic obstacles. Moreover, interaction regulations \cite{Rijksoverheid2017} apply to Amsterdam's canals (\cref{fig:regulations}), similar to those described in the COLlision avoidance REGulations at sea (COLREGs). These regulations are not only mandatory, but adhering to them makes the ASV's motion socially compliant, and therefore, more predictable by other canal users.

\begin{figure}[tp] 
  \centering
  \includegraphics[width=0.45\textwidth]{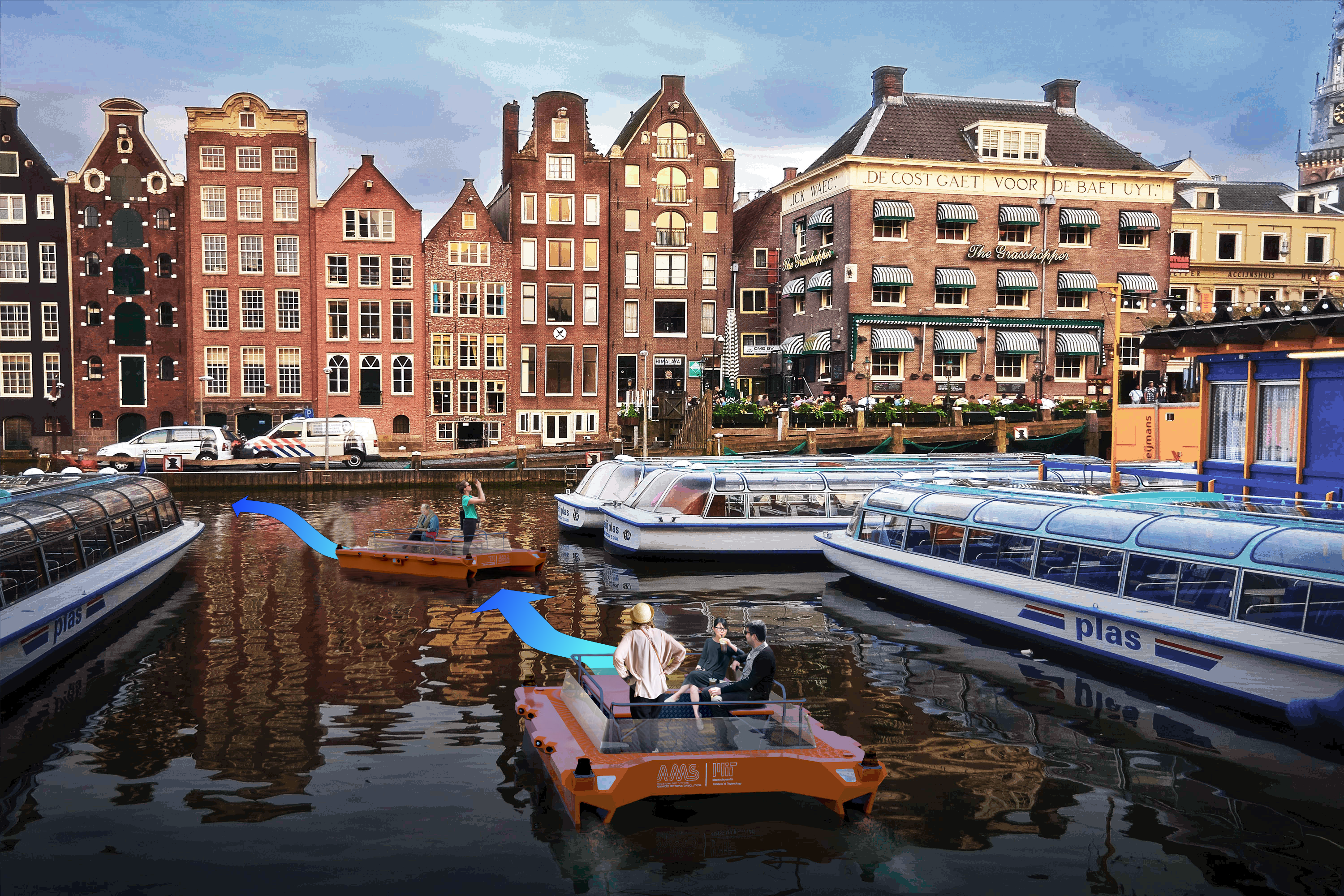}
  \caption{A visualization of two Roboats navigating Amsterdam's canals. The blue arrows represent their planned motion. \textcopyright MIT/AMS Institute.}
  \label{fig:illustration}
\end{figure}

While there are many examples of autonomous cars~\cite{Dixit2020} and mobile robots \cite{Brito2019} navigating in dynamic urban environments containing human agents, the examples for ASVs are limited as they are primarily designed for marine or coastal areas \cite{Liu2016, Gu2020, Zhou2020, Jing2020}. 
Furthermore, motion planning algorithms for autonomous vehicles mostly rely on the road structure \cite{Dixit2020, Brown2017, Liu2017}. On urban waterways there are no traffic lanes or traffic lights, leading to more interactions with other vessels. In order to interact appropriately with obstacle vessels, the motion planning method should be aware of the interaction regulations. Mobile robots often have to deal with similar unstructured dynamic environments \cite{Brito2019, DuToit2012, Kretzschmar2016, Turnwald2019}. However, compared to a mobile robot, a vessel has large inertia and complex dynamics.

\begin{figure}[tp]
  \centering
  \medskip
  \includegraphics[width=\linewidth]{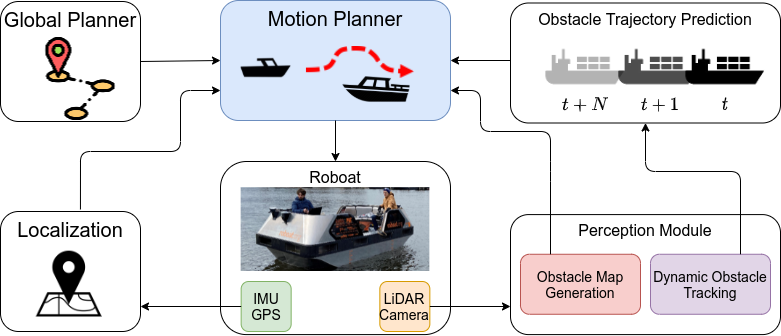}
  \caption{The proposed framework for priority regulations aware motion planning for ASVs in urban canals. The motion planner receives a global path, the current Roboat position, a static obstacle map, and the position and predicted trajectories of the dynamic obstacles. With this information, it generates inputs for Roboat's thrusters.}
  \label{fig:systemoverview}
\end{figure}

In this paper, we propose a motion planning framework for ASVs in urban canals. Our method employs model predictive contouring control (MPCC) \cite{Brito2019} to generate regulation-aware dynamically feasible motions in real-time. The complete system overview is displayed in \cref{fig:systemoverview}. Our main contribution is a method for collision-free and regulations-aware motion planning.

\subsection{Related Work}
Traditional motion planning methods employ a hierarchic planning architecture decomposing the navigation pipeline into a sequence of blocks performing different sub-tasks such as motion planning and control \cite{Paden2016a}. For instance, \cite{Bacha2008} employed A* to search a state lattice and motion primitives for control and \cite{Wang2018} employed A* for path planning and Nonlinear Model Predictive Control (NMPC) as a tracking controller.
However, the first is a reactive method which can be troublesome for collision avoidance in high inertia systems. The second plans along a prediction horizon, but it does not account for static or dynamic obstacles.


Receding-horizon approaches such as Model Predictive Control (MPC) \cite{Brito2019, Dixit2020, Zhu2019, Alcala2020} can be directly deployed in real environments by dealing with the model inaccuracies and environments changes by continuous re-planning online.
However, when navigating in urban canals, ASVs must comply with the inland waterways police regulations \cite{Rijksoverheid2017} which these methods neglect.

COLlision avoidance REGulations at sea (COLREGs) describe the same rules for interactions with other vessels. Several methods implement COLREGs compliance for ASVs in oceanic and coastal environments \cite{Chiang2018, Johansen2016, Eriksen2019, Zhao2016AVessels, Kuwata2014SafeObstacles}. Nevertheless, these approaches are not feasible for a crowded and complex environment like Amsterdam's urban waterways. \cite{Chiang2018} relies on virtual obstacles for COLGRES compliance which makes the problem unfeasible in crowded canals. \cite{Johansen2016} and \cite{Eriksen2019} use a small set of motions primitives that would not be rich enough to navigate dense environments. \cite{Zhao2016AVessels} and \cite{Kuwata2014SafeObstacles} employ geometrical rules resulting in highly reactive motions.
Hence, in this paper, we propose a regulations-aware motion planning framework employing receding-horizon trajectory optimization for static and dynamic collision avoidance.


\subsection{Contribution}
The main contribution is Regulations Aware Model Predictive Contouring Control (RA-MPCC), a motion planning framework for an ASV in urban canals, which includes a cost function that encourages adherence to the interaction regulations in four different scenarios: overtaking, head-on encounter, and crossing with a vessel from starboard or port.

The system is compared in simulation with LMPCC \cite{Brito2019} and the current motion planning and control method for the Roboat, Breadth First Search (BFS) in combination with NMPC \cite{Wang2018} (\cref{sec:simulation}). Moreover, the framework is demonstrated in an outdoor environment with disturbances (\cref{sec:realworldexperiments}).
   
\section{PRELIMINARIES}
Vectors are denoted with bold lowercase symbols, matrices with bold uppercase symbols, and sets with scripted symbols. Superscript $W$ denotes coordinates in World frame, while $B$ indicates the body-fixed frame.

\subsection{Robot Description and Dynamics}
\begin{figure}[tp] 
      \centering
      \smallskip
      \includegraphics[width=0.40\textwidth]{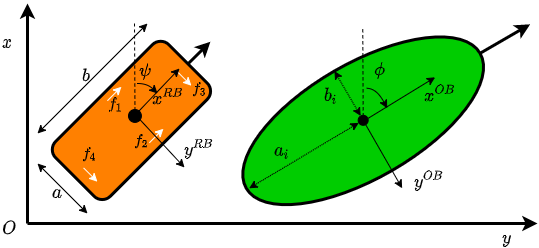}
      \caption{The \textcolor{roboat}{Roboat} and an \textcolor{obstacle}{obstacle vessel}. Angles $\psi$ and $\phi$ denote the orientation of the boats. $a$ and $b$ are the width and the length of the Roboat. $a_i$ and $b_i$ are the semi-major and minor axis respectively of obstacle boat $i$. The Roboat's thrusters can exert force $f$ in two directions, the white arrows denote the positive direction.}
      \label{fig:roboatcoordinates}
   \end{figure}

Let \textit{B} represent an ASV on the plane $\mathcal{W}=\mathbb{R}^2$. The vessel is visualized in \cref{fig:roboatcoordinates}. $\mathbf{p}^W$ denotes the position of the Roboat, and $\mathbf{R}_B^W$ is the rotation matrix corresponding to its orientation. The area occupied by the Roboat is represented with a union of $n_d=3$ discs. Each disc $j$ is centered at a position $\mathbf{p}_j^W = \mathbf{p}^W + \mathbf{R}^W_B(\mathbf{z})\mathbf{p}_j^B$ in the inertial frame, where $\mathbf{p}_j^B$ is the position of the center of disk $j$ in the body frame $B$. 
Moreover, \textit{port} will describe the left side of a boat looking forward, and \textit{starboard} will be used for the right. Additionally, the ASV's dynamics are defined by a discrete-time nonlinear differential equation, as described in \cite{Wang2018}.

\begin{equation}
\begin{split} 
\mathbf{\eta}(t+1) &=\mathbf{R}(\psi)\mathbf{v}(t) \\ 
\mathbf{v}(t+1) &=\mathbf{M}^{-1}\mathbf{Bu}(t)-\mathbf{M}^{-1}(\mathbf{C}(\mathbf{v}(t))+\mathbf{D}(\mathbf{v}(t)))\mathbf{v}(t) \\
\end{split}  
\label{eq:boatdynamics}
\end{equation}

Where the state vector of the vessel is $\mathbf{z}(t) = \left[ \begin{smallmatrix} x & y & \psi & u & v & r \end{smallmatrix} \right]^T$. $\mathbf{\eta}(t) = \left[ \begin{smallmatrix} x & y & \psi \end{smallmatrix} \right]^T$ represents the configuration given by the position and the orientation of the robot in the inertial frame, and $\mathbf{v}(t) = \left[ \begin{smallmatrix} u & v & r \end{smallmatrix} \right]^T$ is respectively the surge velocity, sway velocity, and yaw rate of the vehicle in the body-fixed frame. Converting a state from body frame to inertial frame can be done by the rotation matrix $\mathbf{R}(\psi)$. 
The inputs are given by the four thrusters, the applied forces are described in the control vector $\mathbf{u} = \left[ \begin{smallmatrix} f_1 & f_2 & f_3 & f_4 \end{smallmatrix} \right]^T$. 
The control matrix $\mathbf{B}$ describes the thruster configuration. $\mathbf{M} = \text{diag}\{m_{11},m_{22},m_{33}\}$ is the positive-definitive symmetric added mass and inertia matrix. 
$\mathbf{C(v)} \in \mathbb{R}^{3\times3}$ is the skew-symmetric vehicle matrix of Coriolis and centripetal terms. 
It is assumed that the origin $O_b$ corresponds to the center of mass of the Roboat. $\mathbf{D(v)} \in \mathbb{R}^{3\times3}$ is the positive-semi-definite drag matrix-valued function with linear damping terms on its diagonal. In short, the dynamics can be summarized as:

\begin{equation}
    \mathbf{z}(t+1) = f(\mathbf{z}(t), \mathbf{u}(t))
    \label{eq:simpledynamics}
\end{equation}

\subsection{Static Obstacles}
$\mathcal{O}^\text{static} \subset \mathcal{W}$ is the area that is occupied by static obstacles. This area is represented in an occupancy grid map. This map can either be created beforehand based on map segments of Amsterdam or can be generated in real time from sensor readings.

\subsection{Dynamic Obstacles}
The area occupied by dynamic obstacles, such as boats, is described by $\mathcal{O}^\text{dyn}_k \subset \mathcal{W}$. These dynamic obstacles are represented by ellipsoids with a semi-major axis $a$ and semi-minor axis $b$. For each dynamic obstacle $i$ the current position, rotation matrix $R_i(\phi)$ and velocity $v_i$ are assumed to be known. Future positions of the dynamic obstacles are obtained using a constant velocity model.

\subsection{Global Reference Path}
A global reference path $\mathcal{P}$ is given to our local planner. A global planner could generate this path. The reference is built up from $M$ way-points $p_m^r = [x_m^p, y_m^p] \in \mathcal{W}$ with $m \in \mathcal{M}:=\{1,....,M\}$. As described in \cite{Brito2019}, cubic polynomials describe the path segments for smoothness. A variable $\theta$ represents the traveled distance along the reference path.

\section{MOTION PLANNING}
\label{sec:motionplanning}
This section presents the Regulation Aware Model Predictive Contouring Control (RA-MPCC) method, based on \cite{Brito2019}. This method is used for planning collision-free, dynamically feasible, and regulation-aware motion plans.

At every time step $t$, a receding horizon constrained optimization problem with an $N$ length prediction horizon $T_\text{horizon}$ is solved.

\begin{equation}
\begin{aligned}
    J^* = & \min _{\mathbf{z}_{0:N},\mathbf{u}_{0:N-1},\theta _{0:N}} \sum ^{N-1}_{k =0} J(\mathbf{z}_k, \mathbf{u}_k, \theta _k) + J_N(\mathbf{z}_N,\theta _N)\label{eq:lmpcc}\\ 
    \text{s.t.} \quad & \mathbf{z}_{k+1} = f(\mathbf{z}_{k},\mathbf{u}_{k}), \,\, \theta _{k+1} = \theta _{k} + v_{k} \tau,\\ 
    & \mathcal {B}(\mathbf{z}_k) \cap \big (\mathcal {O}^{\textrm {static}} \cup \mathcal {O}_k^{\textrm {dyn}} \big) = \emptyset,\\ 
    & \mathbf{u}_k \in \mathcal {U}, \,\, \mathbf{z}_{k} \in \mathcal {Z},\, \, \mathbf{z}_{0}, \theta _0 \text { given}.
\end{aligned}
\end{equation}

$J$ is the cost function with $\mathcal{U}$ and $\mathcal{Z}$ the sets of admissible inputs and states and $\mathbf{z}_{0:N}$, $\mathbf{u}_{0:N-1}$ the sequence of state and control inputs, respectively, for the prediction horizon. $\mathcal {B}(\mathbf{z}_k)$ is the space occupied by the Roboat at time-step $k$. The predicted progress along the reference path is $\theta_k$. $v_k$ denotes the forward velocity of the Roboat and $\tau$ is the time-step's length. The output of the optimization is an optimal control input sequence $[\mathbf{u}^*_t]^{t=N-1}_{t=0}$.

\subsection{Cost Function}
The cost function consists of four elements: the tracking, the speed, the input, and the regulations costs (\cref{sec:priorityregulationcompliance}).

\begin{subequations}
\begin{align}
    J_{\text{tracking}}(\mathbf{z}_k,\theta _k) &= \mathbf{e}_k^T \mathbf{Q}_{\epsilon }\mathbf{e}_k \\
    J_{\text{speed}}(\mathbf{z}_k,\mathbf{u}_k) &=Q_v(u_{\text{ref}}-u_k)^2 \\
    J_{\text{input}}(\mathbf{z}_k,\theta_k)&=\mathbf{u}_k^T \mathbf{Q}_u \mathbf{u}_k
\end{align}
\end{subequations}

The tracking cost penalizes error vector $e_k$ containing the estimated contour $\tilde{\epsilon}^c$ and lag $\tilde{\epsilon}^l$ error. 
Second, $J_\text{speed}$ contains the deviation of the surge velocity $u_k$ from the reference velocity $u_\text{ref}$. Furthermore, the inputs are penalized with $J_\text{input}$. $\mathbf{Q}_\epsilon$, $Q_v$, and $\mathbf{Q}_u$ denote design weights. The tracking, velocity, and input costs are further described in \cite{Brito2019}.
The stage cost and the terminal cost are, respectively, $J(\mathbf{z}_k, \mathbf{u}_k, \theta_k):=J_{\text{tracking}} + J_{\text{speed}} + J_{\text{input}} + J_{\text{reg}}$ and $J_N(\mathbf{z}_N, \theta_N) :=J_{\text{tracking}} + J_\text{reg}$, where $J_\text{reg}$ is the regulation cost described in the next section.

\subsection{Priority Regulation Compliance}
\label{sec:priorityregulationcompliance}
In the city of Amsterdam, the inland waterways police regulations \cite{Rijksoverheid2017} apply. Humans operating the other vessels will assume that the Roboat will behave according to these regulations. Thus, complying with these rules will result in social compliance, and therefore, in fewer collisions. 
In this work, we consider the specific regulations that describe interactions with other vessels.
The first interaction regulation describes specific boats to which a small motorboat like the Roboat should always give way. So-called priority boats are sailboats, boats powered by muscle, commercial vessels such as canal cruises, and vessels longer than 20 meters. Still, most of the vessels on the canals are of the same type as the Roboat. When encountering another small motorboat, there are three different situations (\cref{fig:regulations}). First, when having a head-on encounter with another boat, both boats should move to their starboard side. Second, overtaking another boat has to be done, in principle, on their port side. Furthermore, in a crossing, one should give way to a boat from starboard.

   \begin{figure} 
      \centering
      \medskip
      \includegraphics[width=\linewidth]{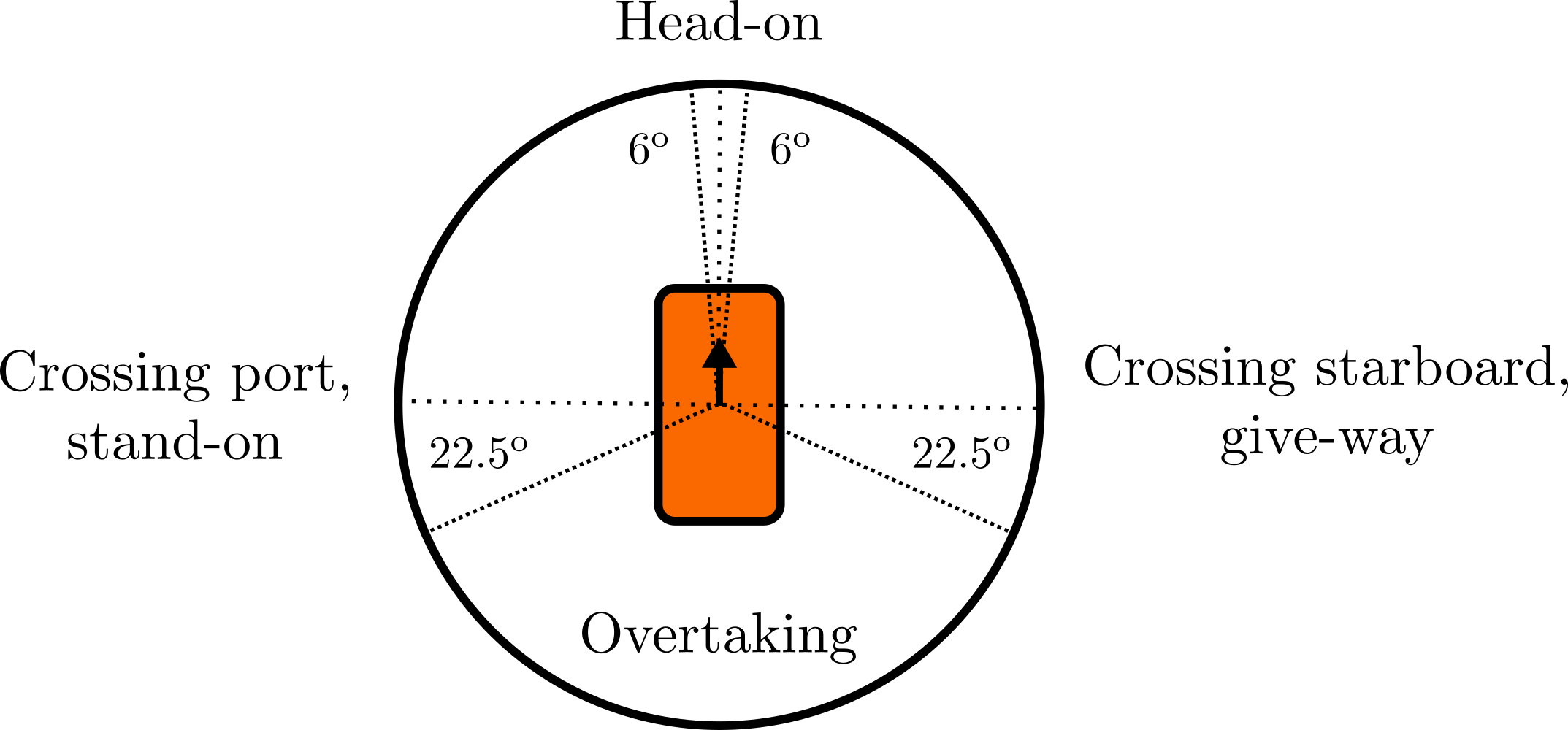}
    \caption{The position of an obstacle boat with respect to the \textcolor{roboat}{Roboat} can be categorized in one of the four regions: head-on, crossing port, crossing starboard or overtaking.}
    \label{fig:foursituationsobstacles}
   \end{figure}

\begin{figure}
    \centering
    \medskip
    \includegraphics[width=0.25\textwidth]{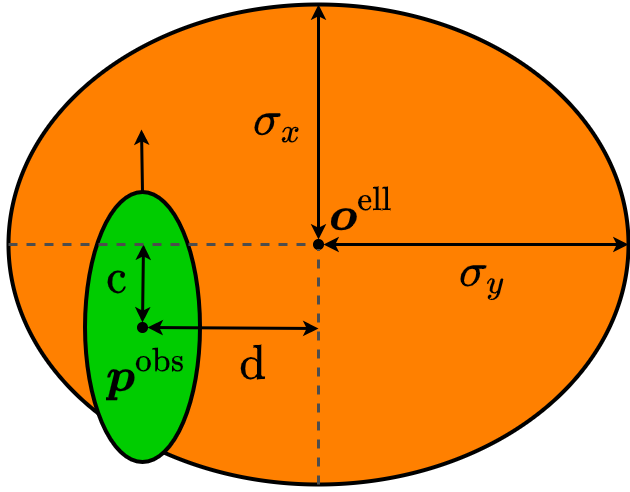}
    \caption{Geometry of the regulations cost function \eqref{eq:2dgaussian}. A higher cost is allocated to the obstacle's starboard and front by shifting the ellipse's center from the center of the obstacle $p^\text{obs}$ by some parameters $c$ and $d$ in the $x$- and $y$-direction of the obstacle, respectively. The standard deviation of the 2D Gaussian for $x$ and $y$ axes are respectively $\sigma_x$ and $\sigma_y$.}
    \label{fig:regulationscosts}
\end{figure}

 While in \cite{Chen2017} and \cite{Eriksen2019}, discrete parameterized cost functions of different shapes were defined with respect to obstacles, a continuous cost function with a simple shape is constructed for our method to facilitate the optimization. We have selected an off-center ellipsoidal 2D Gaussian function \eqref{eq:2dgaussian} as displayed in \cref{fig:regulationscosts}. The idea is to use this smooth function to penalize specific positions with respect to the obstacle boats. We define two types of costs: $J_{HO}$ for Head-on and Overtaking encounters and $J_{RoW}$ for Right of Way (RoW). Both use the same cost function but with different parameters. These two costs together make the regulation costs $J_{reg} = J_{HO} + J_{RoW}$.
 
\subsubsection{Head-On and Overtaking} 
First, the cost function $J_{HO}$ is used to allocate a higher cost to both the starboard side and the front of the obstacle boat. This asymmetric cost will help to achieve the desired trajectories for head-on encounters and overtaking, displayed in \cref{fig:regulations}. The ellipsoid is shifted from the $\mathbf{p}^\text{obs}$ with $c$ in the $x$-direction of the vessel and with $d$ in the negative $y$-direction. This results in the origin of the ellipse $\mathbf{o}^{\text{ell, }W}$ in world frame $W$.

\begin{equation}
\begin{split}
    \mathbf{o}^{\text{ell, }W}_{k, i} &= R(\phi_{k, i}) \begin{bmatrix}c \\ d\end{bmatrix} + \mathbf{p}_{k, i}^\text{obs} = \begin{bmatrix}x_{k, i}^{\text{ell}} \\ y_{k, i}^{\text{ell}} \end{bmatrix}
\end{split}
    \label{eq:centerellipse}
\end{equation}

The standard deviation of the 2D Gaussian ellipsoid in the $x$- and $y$-direction are $\sigma_x$ and $\sigma_y$, respectively. These values are dependent on the size of the obstacle (major-axis $a_i$ and minor-axis $b_i$) and the disc representing the Roboat, and are scaled with parameters $g$ and $h$.

\begin{equation}
\begin{split}
    \sigma_{x, i} &= g (a_i + r_{disc}) \\
    \sigma_{y, i} &= h (b_i + r_{disc})
\end{split}
\label{eq:standarddeviation}
\end{equation}

The cost function $J_{HO}$ can be constructed using scalars $\lambda$, $\mu$ and $\nu$ to shape and rotate the ellipsoid with the obstacle's size and orientation $\phi_i$.

\begin{equation}
\begin{split}
    J_{HO} (\mathbf{z}_k)&=Q_{HO} \sum _{i=1}^n \text{exp} ( - ( \lambda (x^\text{Roboat}_{k} - x^\text{ell}_{k,i })^2\\
    &+ 2 \mu (x^\text{Roboat}_k-x^\text{ell}_{k, i})(y^\text{Roboat}_k-y^\text{ell}_{k, i}) \\
    &+ \nu (y^\text{Roboat}_k-y^\text{ell}_{k, i})^2 )) \\
\end{split}
\label{eq:2dgaussian}
\end{equation}

\begin{equation}
\begin{split}
    \lambda &= \frac{cos(\phi_i)^2}{2\sigma_x^2}+\frac{sin(\phi_i)^2}{2\sigma_y^2} \\
    \mu &= \frac{sin(2\phi_i)}{4\sigma_x^2}-\frac{sin(2 \phi_i)}{4\sigma_y^2}; \\
    \nu &= \frac{sin(\phi_i)^2}{2\sigma_x^2}+\frac{cos(\phi_i)^2}{2\sigma_y^2};
\label{eq:rotationguassian}
\end{split}
\end{equation}

\subsubsection{Right of Way}
The RoW costs will only be allocated to priority vessels depending on their type and length. Additionally, a vessel will also be marked as a priority boat if the Roboat sees the obstacle vessel in the crossing starboard giving way area and the Roboat is seen by the obstacle vessel in the crossing port stand-on area (\cref{fig:foursituationsobstacles}). The cost function will be allocated to the boat for the full prediction horizon.
The cost function is similar to the function for $J_{HO}$, but with different parameters. In this case, a long ellipsoidal cost $J_{RoW}$ weighted with $Q_{RoW}$ will be placed in front of the priority boat to discourage Roboat to stand in its way. For this scenario, the ellipsoid's center will only be shifted in the $x$-direction of the vessel by parameter $f$. Moreover, $\sigma_x$ is equal to parameter $e$, and $\sigma_y$ is $b_i+r_{disc}$.

\subsection{Static Collision Avoidance}
\label{sec:staticcollisionavoidance}
The static obstacles are represented with an occupancy grid map, which is then divided into convex shapes.
After that, the points of these convex shapes that are the closest to the Roboat are selected. At last, the linear constraints are determined such that they are normal to the vector pointing from the Roboat to the closest points. These constraints are defined by a vector $\mathbf{A}$ and a scalar $b$. Resulting in the following equation, in which $\mathbf{p}_j^W$ denotes the position of disc $j$ representing Roboat and $\delta$ is a safety margin.

\begin{equation}
    c^{\text{stat}, j}(\mathbf{z}_0) =  A*\mathbf{p}^W_j - b + r_\text{disc} + \delta \leq 0
    \label{eq:staticconstraint}
\end{equation}


\subsection{Dynamic Collision Avoidance}
For dynamic collision avoidance, it is assumed that the moving obstacles can be represented with an ellipse, having semi-axes $a_i$ and $b_i$. The position and rotation of each obstacle $i \in \{1, ..., n\}$ are $\mathbf{p}_i(t)$ and $\mathbf{R}_i(\phi)$. $\Delta x_{k,i}^j$ and $\Delta y_{k,i}^j$ denote the distance in $x$ and $y$-direction between the center of the disc $j$ and the obstacle $i$. The semi-major axis $\alpha_i = a_i + \delta$ and semi-minor axis $\beta_i = b_i + \delta$ are defined such that any collision will be avoided.
The inequality constraint for each disc of the robot with respect to the obstacle is

\begin{equation}
    c_{k, i}^{\textrm {dyn},j}(\mathbf{z}_{k, i}) = \begin{bmatrix}\Delta x_{k, i}^j\\ \Delta y_{k, i}^j \end{bmatrix}^\textrm {T} R(-\phi_i)^\textrm {T}
    \begingroup 
    \setlength\arraycolsep{2pt}
    \begin{bmatrix}\frac{1}{\alpha_i ^2} & 0\\ 0 & \frac{1}{\beta_i ^2} \end{bmatrix}
    \endgroup
    R(-\phi_i)\ \begin{bmatrix}\Delta x_{k, i}^j\\ \Delta y_{k, i}^j \end{bmatrix} > 1
    \label{eq:dynamicconstraint}
\end{equation}

\section{RESULTS}
\label{sec:results}
This section presents simulation results for different navigation scenarios. First, we introduce the experimental setup used. Then, in \cref{sec:simulation}, we compare our framework with two baseline approaches. We present results demonstrating our method's ability to perform collision avoidance and regulation compliance. Moreover, in \cref{sec:realworldexperiments}, we present experimental results on the MIT's Roboat platform \cite{Wang2018}.

\subsection{Experimental Setup}
\subsubsection{Hardware Setup}
We use the quarter-scale Roboat described in~\cite{Wang2018} for real-world experiments. The entire framework can run on one onboard computer equipped with an Intel i7 CPU. A Velodyne LiDAR is used for perception, and a Swift Nav GPS in combination with a LORD Microstain IMU is used for localization.

\subsubsection{Software Setup}
The motion planner is implemented as a ROS node in C++ and Python. The planner runs onboard at $5$ Hz.
We rely on FORCES PRO \cite{Zanelli2017FORCESProgramsb} for solving the non-convex model predictive contouring control equation \eqref{eq:lmpcc}. If the solver does not find a feasible solution within the maximum number of iterations, a zero command will be sent towards the thrusters, resulting in a deceleration of the Roboat. We use a planning horizon of $10$ s and $20$ steps.
Further, we employ OpenCV \cite{Bradski2000TheLibrary} to divide the occupancy grid into convex shapes for static collision avoidance.

\subsection{Simulation}
\label{sec:simulation}
We compare our approach (RA-MPCC) with the following baseline methods:
\begin{itemize}
    \item \textit{LMPCC} \cite{Brito2019} without regulation awareness. A repulsive cost similar to $J_{HO}$ is still employed, but centered on the obstacle.
    \item \textit{Breadth First Search (BFS) local planner and an NMPC tracking controller} \cite{Wang2018}. Since BFS does not allow for dynamic obstacles in its planning, the space occupied by obstacle boats is added to the static occupancy grid.
\end{itemize}

\begin{figure}[t] 
  \centering
  \medskip
  \includegraphics[width=0.6\linewidth]{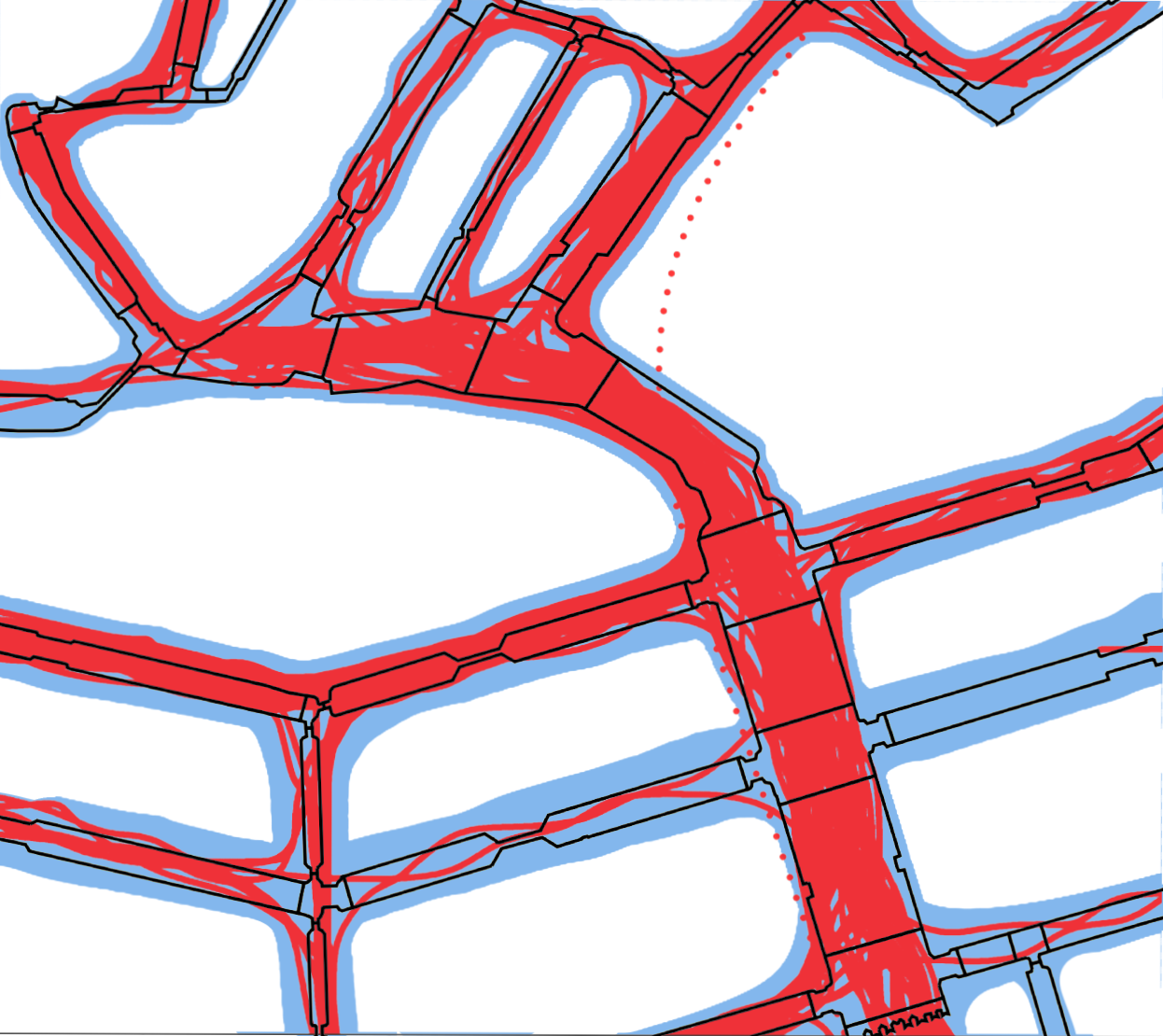}
  \caption{Map of the Amsterdam canals and trajectories collected using the Automatic Identification System (AIS). The canals are depicted in black and in red the vessel's trajectories. Segments of these canals were used as simulation environments.}
  \label{fig:map}
\end{figure}

To simulate the other boats, we replay real vessel trajectories navigating the in the Amsterdam canals, as presented in \cref{fig:map}, collected using the Automatic Identification System (AIS) \cite{AutomaticOverview}. 
The global path is designed such that the Roboat has to interact with the obstacle boat. The eight situations include head-on encounters, taking over other vessels, crossing from starboard, and crossing from port, evenly represented.
Furthermore, we show multi-robot coordination with RA-MPCC in \cref{fig:2irampcc} where two vessels run our method and perform a head-on (left figure) and a crossing  scenario (right figure) while respecting the regulations.

\begin{figure}
    \centering
    \medskip
    \begin{subfigure}[b]{\linewidth}
         \includegraphics[width=\textwidth]{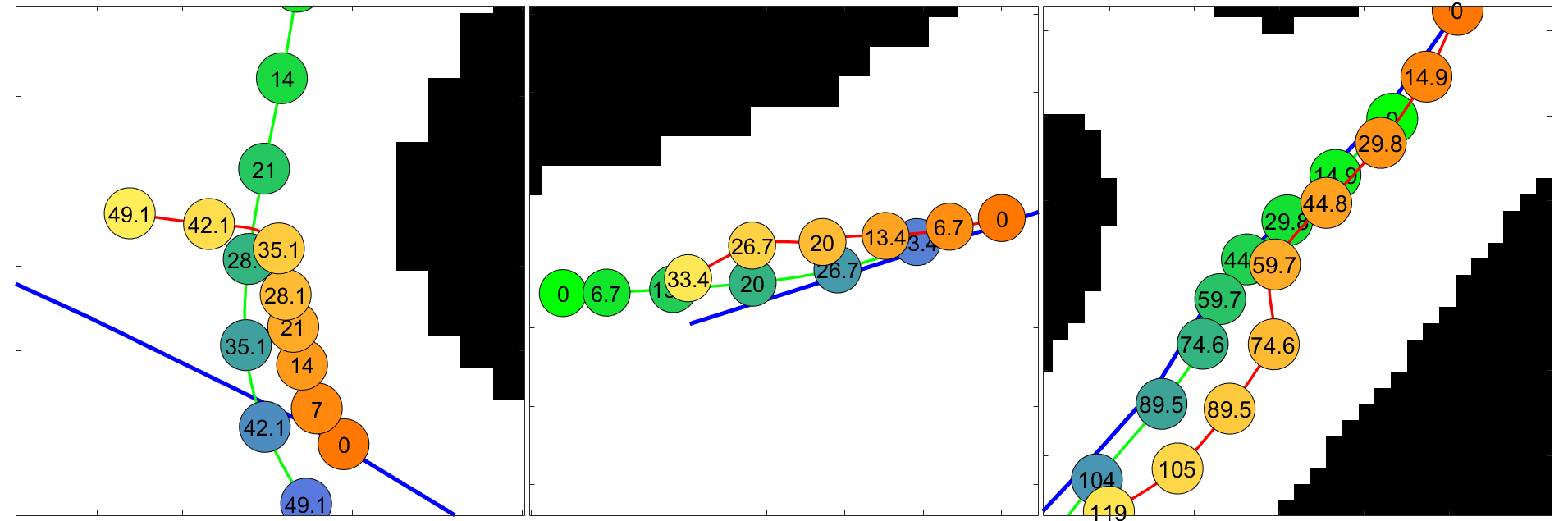}
         \caption{\textit{RA-MPCC} achieves successful obstacle avoidance and compliance to the priority regulations.}
         \label{fig:ourframeworkconstvel}
    \end{subfigure}
    \begin{subfigure}[b]{\linewidth}
        \includegraphics[width=\textwidth]{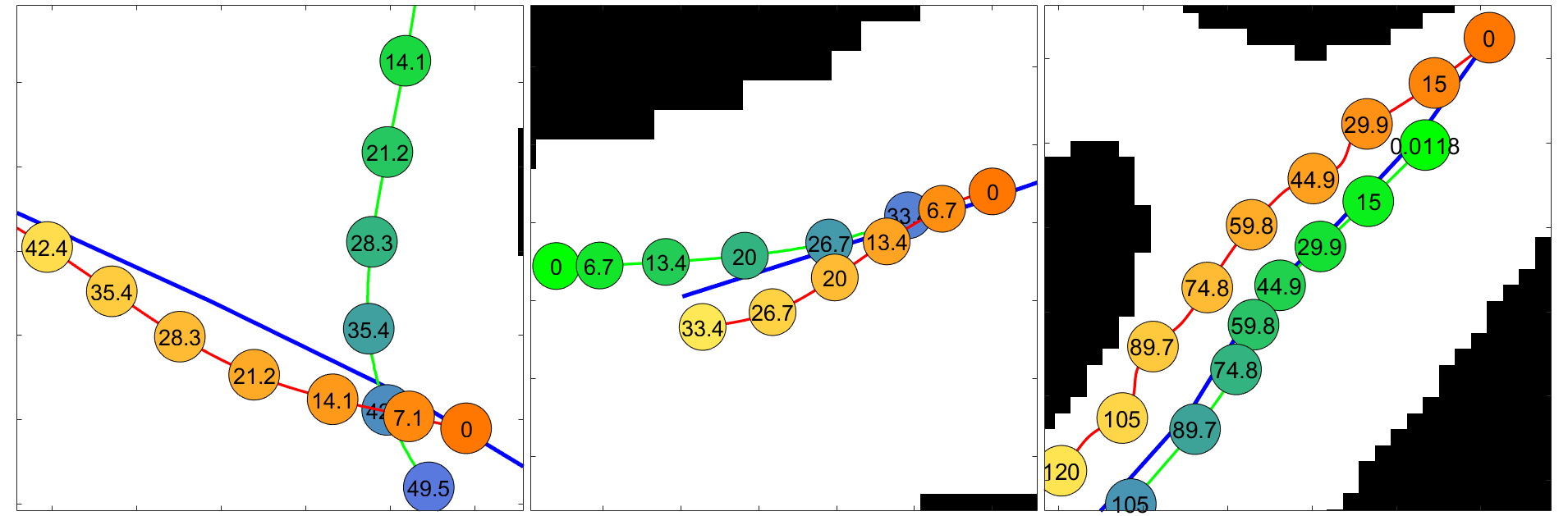}
         \caption{\textit{LMPCC} also results in successful obstacle avoidance but does not generate regulation-aware trajectories.}
         \label{fig:ourframeworkcentralcosts}
    \end{subfigure}
    \begin{subfigure}[b]{\linewidth}
        \includegraphics[width=\textwidth]{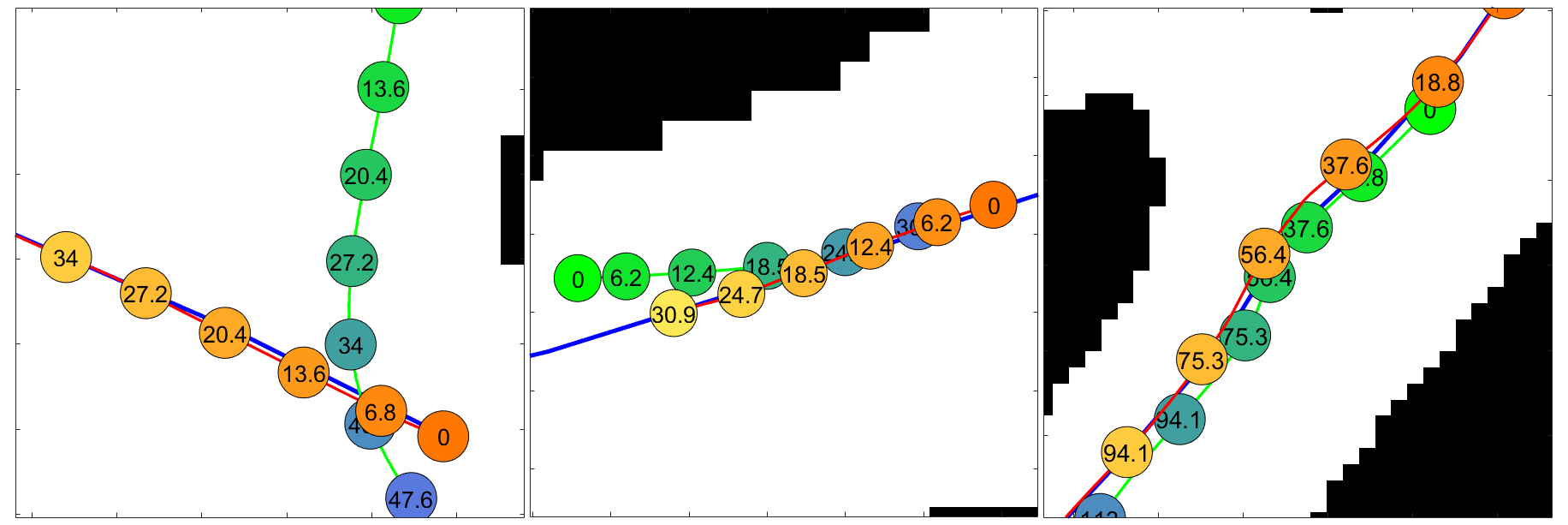}
         \caption{\textit{BFS combined with NMPC} follows the global path closely and is not able to anticipate the moving obstacles. We see more dangerous situations and no regulation compliance.}
         \label{fig:bfsandnmpc}
    \end{subfigure}
    \caption{The Roboat (\textcolor{roboat}{orange}-\textcolor{roboatyellow}{yellow}) avoiding an obstacle boat (\textcolor{obstacle}{green}-\textcolor{obstacleblue}{blue}) while following a \textcolor{refpath}{global reference path}. The obstacle boat is executing a trajectory taken from a real-world data set. Timestamps are displayed in seconds, and the static obstacles are represented in black.}
    \label{fig:comparisonsimulation}
\end{figure}


\subsubsection{Dynamic Collision Avoidance}
The results presented in \cref{tab:resultssimulation} demonstrate that our approach outperforms the baseline methods in terms of percentage of collisions. Qualitative results presented in \cref{fig:comparisonsimulation}, shows that RA-MPCC and the original LMPCC method can avoid dynamic obstacles in different situations. However, the head-on and take over cost function $J_{HO}$ can help the solver choose between avoidance on the starboard or the port side, therefore starting the manoeuvre earlier on and resulting in a safer motion.
In contrast, the BFS combined with the NMPC closely follows the global path without considering the future obstacle positions, resulting in a high number of collisions.

\subsubsection{Compliance to Regulations}
\begin{figure}
    \centering
    \includegraphics[width=0.4\textwidth]{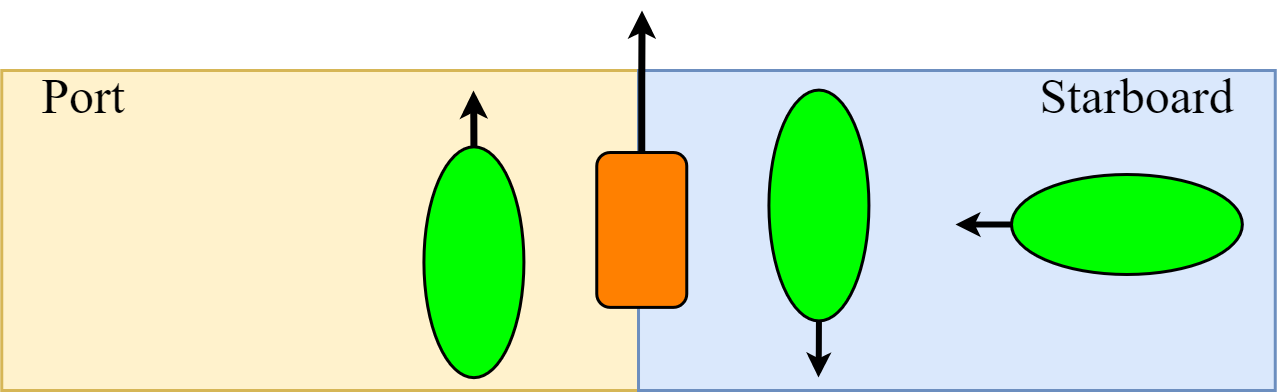}
    \caption{Violations of the right-handed priority regulations. The \textcolor{roboat}{Roboat} should never be in a situation where an \textcolor{obstacle}{obstacle boat} is in one of the configurations shown in the figure.}
    \label{fig:regulationviolations}
\end{figure}

Similar to \cite{Chen2017}, we define situations where the regulations are breached for both the right-handed and the left-handed rules. Right-handed regulations are the norm for vessels, for example, giving way to someone from starboard. In contrast, left-handed regulations require giving way to a vessel nearing from port. Violations to these regulations are registered when an obstacle vessel is in one of the rectangular regions with a specific orientation with respect to the Roboat (see \cref{fig:regulationviolations}) for more than $dt$ seconds. For this experiment we have used $dt = 0.17$. 

\begin{table}
\begin{tabular}[width=0.8\linewidth]{ccccc}
\textbf{Method}  & \textbf{\% Right-Handed Violations } & \textbf{\% Collisions }            \\ \hline
\textbf{LMPCC} & 64.20  & 1.23           \\ \hline
\textbf{BFS + NMPC}  & 68.32  & 36.36             \\ \hline
\textbf{RA-MPCC}   & \textbf{19.38}  & \textbf{0.00}    \\ \hline 
\end{tabular}
\caption{Results for eight different scenarios, each ran ten times. These test cases include head-on, overtaking, crossing from port and starboard. Out of all violations (left and right-handed), the percentage of right-handed ones is calculated for each run. The percentage shows the mean over all runs.}
\label{tab:resultssimulation}
\end{table}

\begin{figure}[ht]
    \centering
    \medskip
     \includegraphics[width=0.40\textwidth]{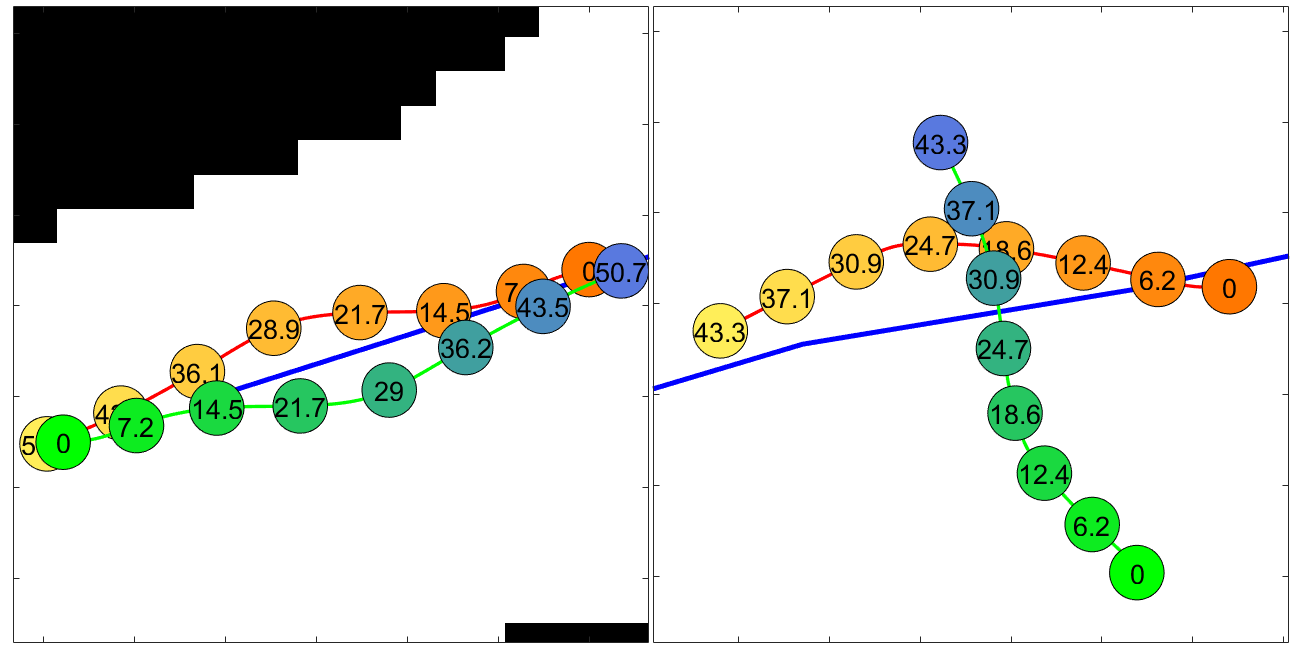}
    \caption{Two vessels both running the proposed RA-MPCC method. On the left, the two vessels encounter each other head-on. On the right, a crossing is performed. In both scenarios the regulations are satisfied. Timestamps are displayed in seconds.}
    \label{fig:2irampcc}
\end{figure}

The quantitative results presented in \cref{tab:resultssimulation} and qualitative results displayed in  \cref{fig:comparisonsimulation}, show that RA-MPCC incurs in the lower number of right-handed violations relative to the left-handed ones. RA-MPCC mainly produces paths that pass the other boat on their port side when facing head-on encounters or overtaking. LMPCC and BFS, instead, are more likely to cross with the obstacle vessel's starboard side, not complying with the inland waterway regulations. Moreover, our method will plan a trajectory that does not cross the trajectory of an obstacle approaching from starboard. Furthermore, \cref{fig:2irampcc} shows two decentralized multi-robot coordination scenarios where all agents ran our RA-MPCC without communicating with each other. In both the head-on and crossing situations, the vessels avoid each other while complying with the regulations.

\subsection{Real-World Experiments}
\label{sec:realworldexperiments}
\begin{figure}[tp] 
  \centering
  \includegraphics[width=0.8\linewidth]{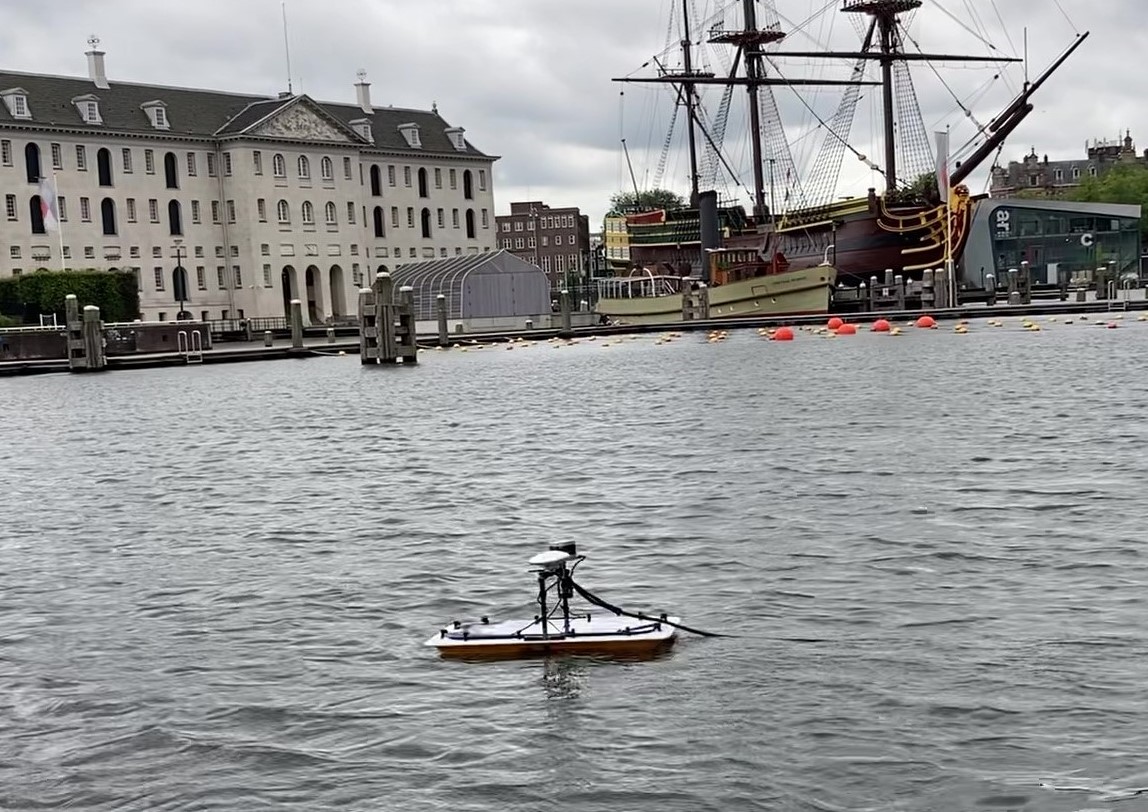}
  \caption{The quarter-scale Roboat \cite{Wang2018} using RA-MPCC at the Marineterrein in Amsterdam.}
  \label{fig:marineterrein}
\end{figure}

RA-MPCC was implemented on a quarter-scale Roboat \cite{Wang2018} for testing at the Marineterrein in Amsterdam (\cref{fig:marineterrein}). For more details on the real-world experiments, we refer the reader to the accompanying video.



\section{CONCLUSIONS AND FUTURE WORK}
This paper proposed a motion planning framework called RA-MPCC for ASVs in urban canals based on LMPCC. This framework is able to plan local trajectories that avoid dynamic obstacles according to the regulations. We compare our method to the original LMPCC method and a BFS local planner combined with an NMPC tracking controller. Simulated experiments, executed on Amsterdam's canal segments with real vessel trajectory data, have shown that RA-MPCC outperforms both methods in urban canals. Moreover, we have shown that RA-MPCC also performs well in a two-agent coordination scenario where all vessels run the proposed method. 
As future work, better predictions of the obstacle vessels' motion can be incorporated into the framework to plan more efficient trajectories in crowded environments.





\section*{ACKNOWLEDGMENT}
We would like to acknowledge the staff from the Roboat project at AMS Institute for assisting with the experimental tests. In particular, we would like to thank Jonathan Klein Schiphorst, Joshua Jordan, and Rens Doornbusch. In addition, we would like to acknowledge the MIT team for providing the quarter-scale Roboat and the NMPC code.

\bibliography{references.bib}
\bibliographystyle{IEEEtran}

\end{document}